\documentclass[11pt]{article}

\usepackage[final]{acl}

\usepackage{times}
\usepackage{latexsym}
\usepackage[T1]{fontenc}
\usepackage[utf8]{inputenc}
\usepackage{microtype}
\usepackage{inconsolata}
\usepackage{comment}
\usepackage{amsmath,amssymb,amsfonts}
\usepackage{graphicx}
\usepackage{wrapfig}
\usepackage{booktabs}
\usepackage{subcaption}
\usepackage{textcomp}
\usepackage{booktabs}
\usepackage{multirow}
\usepackage{makecell}
\usepackage{float}
\usepackage{tabularx}
\usepackage{array}

\usepackage{url}
\usepackage{hyperref}
\usepackage{float}
\usepackage{algorithm}
\usepackage{algorithmic}
\usepackage{listings}
\usepackage[table]{xcolor}

\graphicspath{{./images/}{./Images/}}

\newcommand{\PreserveBackslash}[1]{\let\temp=\\#1\let\\=\temp}
\newcolumntype{C}[1]{>{\PreserveBackslash\centering}p{#1}}
\newcolumntype{R}[1]{>{\PreserveBackslash\raggedleft}p{#1}}
\newcolumntype{L}[1]{>{\PreserveBackslash\raggedright}p{#1}}

\lstdefinelanguage{json}{
    basicstyle=\scriptsize\ttfamily,
    showstringspaces=false,
    breaklines=true,
    frame=lines,
    xleftmargin=5pt,
    xrightmargin=5pt,
    stringstyle=\color{teal},
    commentstyle=\color{gray},
    morestring=[b]",
    literate=
     *{0}{{{\color{black}0}}}{1}
      {1}{{{\color{black}1}}}{1}
      {2}{{{\color{black}2}}}{1}
      {3}{{{\color{black}3}}}{1}
      {4}{{{\color{black}4}}}{1}
      {5}{{{\color{black}5}}}{1}
      {6}{{{\color{black}6}}}{1}
      {7}{{{\color{black}7}}}{1}
      {8}{{{\color{black}8}}}{1}
      {9}{{{\color{black}9}}}{1},
    keywordstyle=\color{blue},
    moredelim=[s][\color{blue}]{\"}{\":},
}

\title{MemeCULT-1K: Benchmarking South Asian Cultural Context and Humor Understanding of Multimodal Models}

\author{
  \begin{tabular}{c}
    Tawsif Tashwar Dipto\textsuperscript{1}\thanks{Equal contribution.\quad\textsuperscript{\textdagger}Equal supervision.}\quad
    Mehedi Ahamed\textsuperscript{2}\footnotemark[\value{footnote}]\quad
    Radib Bin Kabir\textsuperscript{1}\footnotemark[\value{footnote}] \\[4pt]
    Mueeze Al Mushabbir\textsuperscript{1}\quad
    Mohammed Saidul Islam\textsuperscript{3}\quad
    Mir Rayat Imtiaz Hossain\textsuperscript{4} \\[4pt]
    Md Tahmid Rahman Laskar\textsuperscript{5}\textsuperscript{\textdagger}\quad
    Sabbir Ahmed\textsuperscript{1,6}\textsuperscript{\textdagger} \\[7pt]
    \normalfont\textsuperscript{1}Islamic University of Technology\quad
    \textsuperscript{2}South East University \quad \textsuperscript{3}Vector Institute\\[2pt]
    \normalfont\textsuperscript{4}University of British Columbia\quad \textsuperscript{5}York University \quad \textsuperscript{6}Queen's University \\[6pt]
    \normalfont\texttt{\{tawsiftashwar, radib\}@iut-dhaka.edu}\quad
    \normalfont\texttt{mehedi.ahamed@seu.edu.bd}
  \end{tabular}
}

\begin{document}
\maketitle

\begin{abstract}
Meme understanding goes beyond recognizing visual content or literal text; it
requires implicit cultural knowledge and pragmatic inference that most
vision-language models still lack. We introduce \textbf{MemeCULT-1K}, a
multilingual benchmark of 1,000 South Asian memes in Bengali, English, and
Hindi, where each meme is paired with a cultural context note and three
human-written explanations, along with a supplementary set of 54 Bengali
regional dialect memes. We evaluate thirteen popular Vision Language Models
(VLMs) under two settings: \emph{meme-only} and \emph{context-aware}.
Providing minimal cultural context yields consistent gains across all models and languages: mean SBERT similarity improves from 44.6 to 56.4 ($+$11.8), BLEURT from 37.3 to 42.3 ($+$5.0), and LLM-as-a-Judge scores from 2.57 to 3.43 out of 5 ($+$0.86).
Fine-grained error analysis reveals that closed-source models fail mainly on
entity and reference misidentification, while open-source models are
bottlenecked by broader cultural knowledge gaps, with linguistic and
phonological failures proving the most context-resistant across both. These
results highlight the difficulty of culturally grounded meme understanding and
motivate future work on explicit cultural knowledge integration. Our dataset and code are publicly available at \href{https://github.com/TawsifDipto17/MemeCULT-1K}{TawsifDipto17/MemeCULT-1K}.

\end{abstract}

\section{Introduction}

Memes are multimodal forms of online humor whose meaning often depends on cultural knowledge, social context, and pragmatic inference rather than on image or text recognition alone \citep{nguyen2024computational}. A model may correctly parse both the caption and visual content, yet still fail to understand the humor when the underlying cultural reference, person, event, idiom, or social situation is not recognized \citep{ananthram2025see}. This makes meme understanding a challenging test of culturally grounded vision-language reasoning.

Existing meme benchmarks largely focus on classification tasks such as sentiment, offensiveness, or hatefulness \citep{sharma2020semeval,kiela2020hateful}, while explanation-centric resources such as MemeCap \citep{hwang-shwartz-2023-memecap} and MEMEX \citep{sharma-etal-2023-memex} remain English-dominant and Western-centric \citep{nguyen2024computational}. South Asian memes provide a particularly challenging setting because they frequently combine Bengali, Hindi, English, and code-mixed text with references to cricket, cinema
and region-specific humor conventions.
Since these references are often hard to infer from the image and caption alone, in this paper, we propose a context-aware evaluation setting that tests whether model failures stem from missing background knowledge.

To support this investigation, we introduce \textsc{MemeCULT-1K}, a benchmark
of over 1,000 South Asian memes in Bengali, English, and Hindi. Each meme is paired
with a short cultural context note and three independent human-written English
explanations. We evaluate thirteen closed and open-source VLMs under two
settings: \emph{meme-only}, where the model sees only the meme, and
\emph{context-aware}, where it also receives the cultural context note. We
additionally include 54 Bengali regional dialect memes to probe model robustness
to non-standard spellings, dialectal vocabulary, and localized humor conventions. Across automatic metrics, LLM-as-a-judge scoring, and human evaluation, context consistently improves explanation quality. Our error analysis identifies weaknesses in both open-source and closed-source models for cultural contextual meme understanding.

\begin{figure*}[]
\centering
\includegraphics[width=0.90\linewidth]{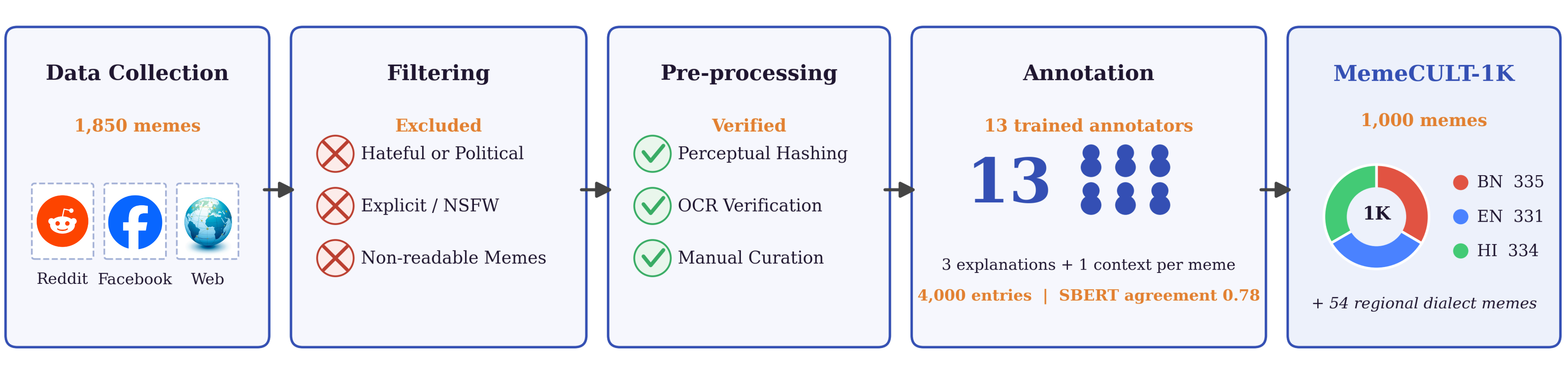}
\caption{\small{Dataset pipeline: sourcing, filtering, cleaning, and annotation with validation to produce the \textsc{MemeCULT-1K} dataset.}}

\label{fig:dataset_creation}
\vspace{-3mm}
\end{figure*}

Overall, our contributions are threefold: (i) \textsc{MemeCULT-1K}, a multilingual benchmark for South Asian meme explanation; (ii) a context-aware evaluation setting for measuring the cultural background in VLM reasoning; and (iii) a fine-grained analysis of persistent VLM failures in culturally grounded meme understanding.

\section{Related Work}
Existing meme benchmarks mainly frame meme understanding as a \textit{classification} task, covering sentiment, humor labels, hatefulness, misogyny, and language-specific abusive content \cite{sharma2020semeval,kiela2020hateful,GASPARINI2022108526,das-mukherjee-2023-banglaabusememe}. While useful, these classification labels do not directly assess whether a system can recover the \emph{implicit premise} that makes a meme meaningful and funny.

Explanation-centric meme evaluation better captures interpretation, with
datasets such as MemeCap \cite{hwang-shwartz-2023-memecap} and MEMEX
\cite{sharma-etal-2023-memex} pairing English memes with captions or
explanatory evidence, but such resources remain English- and Western-centric
\cite{nguyen2024computational}. More broadly, although VLMs perform strongly
on standard multimodal benchmarks \cite{li2022blip,radford2021clip,li2023blip2},
prior work shows that they degrade under real-world ambiguity
\cite{lee2025visionwild} and in culturally specific settings, often missing
culturally grounded intent despite fluent descriptions \cite{ananthram2025see}.
These gaps motivate our focus on culturally grounded South Asian meme
explanation with explicit context-aware evaluation.

\definecolor{eng}{RGB}{31,119,180}   
\definecolor{ben}{RGB}{214,39,40}    
\definecolor{hin}{RGB}{44,160,44}    

\newcommand{\en}[1]{\textcolor{eng}{#1}}
\newcommand{\bn}[1]{\textcolor{ben}{#1}}
\newcommand{\hi}[1]{\textcolor{hin}{#1}}
\newcommand{\langsep}{\;\;|\;\;}

\begin{table*}[t]
\centering
\small
\renewcommand{\arraystretch}{1.08}
\setlength{\tabcolsep}{2pt}
\resizebox{\linewidth}{!}{
\begin{tabular}{lcc|c|cc|c|cc|c}
\toprule

\textbf{Model} 
& \multicolumn{2}{c|}{\textbf{SBERT Similarity}} 
& \textbf{Avg. $\Delta$}
& \multicolumn{2}{c|}{\textbf{BERTScore F1}} 
& \textbf{Avg. $\Delta$}
& \multicolumn{2}{c|}{\textbf{BLEURT}} 
& \textbf{Avg. $\Delta$} \\
\cmidrule(lr){2-3} 
\cmidrule(lr){5-6} 
\cmidrule(lr){8-9}

& \textbf{Meme-only} & \textbf{Context-aware} 
&& \textbf{Meme-only} & \textbf{Context-aware} 
&& \textbf{Meme-only} & \textbf{Context-aware} 
& \\
\midrule

GPT-5 Nano 
& \en{48.1}\langsep\bn{36.8}\langsep\hi{49.3} 
& \en{54.8}\langsep\bn{55.3}\langsep\hi{52.7}
& +9.53
& \en{85.6}\langsep\bn{84.3}\langsep\hi{85.1} 
& \en{86.2}\langsep\bn{85.7}\langsep\hi{85.5}
& +0.80
& \en{39.1}\langsep\bn{33.1}\langsep\hi{38.7} 
& \en{41.2}\langsep\bn{40.7}\langsep\hi{40.6}
& +3.87 \\

GPT-5 Mini 
& \en{51.9}\langsep\bn{49.7}\langsep\hi{51.3} 
& \en{56.0}\langsep\bn{58.8}\langsep\hi{54.4}
& +5.43
& \en{85.7}\langsep\bn{85.1}\langsep\hi{85.3} 
& \en{86.1}\langsep\bn{85.9}\langsep\hi{85.5}
& +0.47
& \en{40.2}\langsep\bn{39.1}\langsep\hi{39.7} 
& \en{41.3}\langsep\bn{42.3}\langsep\hi{41.1}
& +1.90 \\

Gemini 2.5 Flash Lite 
& \en{57.9}\langsep\bn{51.8}\langsep\hi{56.7} 
& \en{59.6}\langsep\bn{59.1}\langsep\hi{58.7}
& +3.67
& \en{86.5}\langsep\bn{85.5}\langsep\hi{86.0} 
& \en{86.9}\langsep\bn{86.5}\langsep\hi{86.3}
& +0.57
& \en{41.0}\langsep\bn{38.7}\langsep\hi{41.5} 
& \en{42.9}\langsep\bn{41.1}\langsep\hi{42.9}
& +1.90 \\

Gemini 2.5 Flash 
& \en{55.8}\langsep\bn{54.1}\langsep\hi{53.8} 
& \en{58.2}\langsep\bn{56.6}\langsep\hi{54.9}
& +2.00
& \en{86.5}\langsep\bn{86.0}\langsep\hi{85.9} 
& \en{86.8}\langsep\bn{86.3}\langsep\hi{86.0}
& +0.23
& \en{41.8}\langsep\bn{41.1}\langsep\hi{40.9} 
& \en{43.2}\langsep\bn{42.6}\langsep\hi{41.9}
& +1.30 \\

Gemini 2.5 Pro
& \en{57.1}\langsep\bn{56.9}\langsep\hi{54.2} 
& \en{59.4}\langsep\bn{58.6}\langsep\hi{54.7}
& +1.50
& \en{86.9}\langsep\bn{86.7}\langsep\hi{86.2} 
& \en{87.3}\langsep\bn{86.9}\langsep\hi{86.4}
& +0.27
& \en{43.7}\langsep\bn{44.0}\langsep\hi{42.1} 
& \en{45.0}\langsep\bn{45.2}\langsep\hi{42.3}
& +0.90 \\

\midrule

Gemma-3-4B 
& \en{31.6}\langsep\bn{22.9}\langsep\hi{31.4} 
& \en{37.6}\langsep\bn{36.0}\langsep\hi{36.9}
& +8.20
& \en{85.6}\langsep\bn{84.9}\langsep\hi{85.4} 
& \en{85.8}\langsep\bn{85.4}\langsep\hi{85.5}
& +0.27
& \en{33.9}\langsep\bn{30.6}\langsep\hi{34.3} 
& \en{36.1}\langsep\bn{35.5}\langsep\hi{36.4}
& +3.07 \\

Gemma-3-12B 
& \en{37.0}\langsep\bn{24.8}\langsep\hi{34.4} 
& \en{52.9}\langsep\bn{49.5}\langsep\hi{49.9}
& +18.70
& \en{85.8}\langsep\bn{84.7}\langsep\hi{85.5} 
& \en{86.7}\langsep\bn{85.9}\langsep\hi{86.0}
& +0.87
& \en{34.6}\langsep\bn{28.4}\langsep\hi{33.4} 
& \en{40.3}\langsep\bn{38.0}\langsep\hi{39.1}
& +7.00 \\

\midrule

InternVL3.5-2B 
& \en{49.4}\langsep\bn{25.3}\langsep\hi{44.0} 
& \en{66.1}\langsep\bn{62.2}\langsep\hi{62.5}
& +24.03
& \en{86.6}\langsep\bn{84.9}\langsep\hi{85.3} 
& \en{88.0}\langsep\bn{87.2}\langsep\hi{86.8}
& +1.73
& \en{38.5}\langsep\bn{31.0}\langsep\hi{37.2} 
& \en{46.7}\langsep\bn{44.3}\langsep\hi{44.4}
& +9.57 \\

InternVL3.5-4B 
& \en{53.7}\langsep\bn{28.6}\langsep\hi{49.4} 
& \en{67.7}\langsep\bn{63.4}\langsep\hi{63.3}
& +20.90
& \en{86.9}\langsep\bn{84.9}\langsep\hi{85.3} 
& \en{88.2}\langsep\bn{87.2}\langsep\hi{87.8}
& +2.03
& \en{40.0}\langsep\bn{31.2}\langsep\hi{38.6} 
& \en{46.6}\langsep\bn{44.0}\langsep\hi{44.3}
& +8.37 \\

InternVL3.5-8B 
& \en{52.7}\langsep\bn{30.8}\langsep\hi{51.4} 
& \en{61.7}\langsep\bn{58.2}\langsep\hi{57.7}
& +14.23
& \en{86.7}\langsep\bn{85.0}\langsep\hi{83.8} 
& \en{87.5}\langsep\bn{86.8}\langsep\hi{86.9}
& +1.90
& \en{40.2}\langsep\bn{31.1}\langsep\hi{39.7} 
& \en{44.4}\langsep\bn{42.3}\langsep\hi{43.0}
& +6.23 \\

\midrule

Qwen3VL-2B 
& \en{53.6}\langsep\bn{31.7}\langsep\hi{46.8} 
& \en{67.7}\langsep\bn{62.8}\langsep\hi{61.6}
& +20.00
& \en{86.4}\langsep\bn{84.3}\langsep\hi{84.0} 
& \en{87.8}\langsep\bn{87.0}\langsep\hi{85.0}
& +1.70
& \en{39.8}\langsep\bn{32.3}\langsep\hi{37.4} 
& \en{46.7}\langsep\bn{43.9}\langsep\hi{44.2}
& +8.43 \\

Qwen3VL-4B 
& \en{47.9}\langsep\bn{28.0}\langsep\hi{45.3} 
& \en{56.4}\langsep\bn{55.5}\langsep\hi{53.8}
& +14.83
& \en{86.0}\langsep\bn{84.7}\langsep\hi{83.9} 
& \en{86.9}\langsep\bn{86.4}\langsep\hi{86.4}
& +1.70
& \en{37.9}\langsep\bn{30.3}\langsep\hi{36.8} 
& \en{42.3}\langsep\bn{41.4}\langsep\hi{41.3}
& +6.67 \\

Qwen3VL-8B 
& \en{50.2}\langsep\bn{34.1}\langsep\hi{49.5} 
& \en{57.0}\langsep\bn{56.8}\langsep\hi{49.7}
& +9.90
& \en{86.2}\langsep\bn{84.9}\langsep\hi{84.1} 
& \en{86.9}\langsep\bn{86.4}\langsep\hi{88.3}
& +2.13
& \en{39.0}\langsep\bn{32.0}\langsep\hi{38.2} 
& \en{42.4}\langsep\bn{41.3}\langsep\hi{40.8}
& +5.10 \\

\bottomrule
\end{tabular}}
\caption{\small{Automatic metric results on \textsc{MemeCULT-1K}. Each entry reports \en{English} \langsep \bn{Bengali} \langsep \hi{Hindi}. Avg. $\Delta$ denotes the average change in performance after providing context across all three languages. Higher is better.}}
\label{tab:quantitative_results}
\end{table*}

\section{Dataset}
\label{sec:dataset}
We introduce \textsc{MemeCULT-1K}, a multilingual dataset of 1,000 South Asian
memes. The dataset contains \textbf{Bengali (335), English (331), and Hindi (334)} memes,
with frequent code-mixing and culturally specific references. We additionally
collected 54 Bengali regional dialect memes as a supplementary evaluation set.
Each meme is paired with three independent English explanations and one short
cultural context note. The dataset covers diverse domains, including sports,
education, pop culture, irony, and social commentary. Figure~\ref{fig:dataset_creation} summarizes our data creation pipeline.

\noindent \textbf{Data Collection:}
1,850 memes were collected through manual curation and web scraping from publicly accessible sources, such as Reddit, Facebook, and regional meme pages. 
We excluded explicit, hateful, and political content. 

\noindent \textbf{Pre-processing:}
From 1,850 collected memes, we removed duplicates and low-quality images using perceptual hashing and manual review. OCR was used to verify language accuracy, and illegible memes were discarded. Three curators further checked cultural relevance, including code-mixed Banglish and Hinglish examples (a mix of English and Bengali). The final dataset contains 1,000 balanced memes.

\noindent \textbf{Annotation:}
Thirteen trained annotators, fluent in English and at least in one target language, wrote three English explanations and one cultural context note for each meme, yielding 4,000 text entries. 
To ensure quality, annotations were validated through majority agreement, with curators stepping in to resolve any disagreements. All contributors were expected to write in a neutral tone, with proper grammar, and with cultural sensitivity in mind. The mean pairwise SBERT similarity \cite{reimers2019sentencebert} among explanations was 0.78, indicating strong semantic agreement.

\noindent \textbf{Difficulty Stratification:}
To characterize the sources of difficulty across languages, we annotate all 1,000 memes with a dominant difficulty type, four non-exclusive properties (wordplay, code-mixing, cultural prior knowledge, and entity recognition), and reference scope. As shown in Table~\ref{tab:difficulty_stratification}, situational or relatable context is the most common difficulty overall (52.4\%), particularly in Bengali (71.6\%) and Hindi (58.7\%), while English exhibits a more balanced distribution across situational, cultural/social, and linguistic difficulties. Cultural prior knowledge is required for 33.9\% of memes and entity recognition for 17.1\%, with substantial variation across languages; references are predominantly national (51.0\%) or international (47.7\%). Overall, these results highlight that multilingual meme understanding depends on diverse forms of contextual knowledge beyond visual and textual content, whose relative importance varies considerably across languages.

\begin{table}[t]
\centering
\resizebox{0.8\columnwidth}{!}{%
\begin{tabular}{lrrrr}
\toprule
\textbf{Meme property}
& \en{\textbf{En}}
& \bn{\textbf{Bn}}
& \hi{\textbf{Hi}}
& \textbf{All} \\
\midrule

\multicolumn{5}{l}{\textbf{Dominant difficulty type}} \\
Situational / Relatable
& \en{26.6} & \bn{71.6} & \hi{58.7} & 52.4 \\
Entity / Reference
& \en{19.3} & \bn{5.4} & \hi{19.8} & 14.8 \\
Cultural / Social knowledge
& \en{26.9} & \bn{13.4} & \hi{16.5} & 18.9 \\
Linguistic / Code-mixing
& \en{20.5} & \bn{2.4} & \hi{0.9} & 7.9 \\
Wordplay / Pun
& \en{6.6} & \bn{7.2} & \hi{4.2} & 6.0 \\
\midrule

\multicolumn{5}{l}{\textbf{Property present (\% Yes)}} \\
Pun / wordplay
& \en{7.3} & \bn{8.4} & \hi{4.2} & 6.6 \\
Code-mixing
& \en{39.6} & \bn{5.4} & \hi{2.7} & 15.8 \\
Cultural prior needed
& \en{41.1} & \bn{31.3} & \hi{29.3} & 33.9 \\
Entity recognition needed
& \en{21.5} & \bn{9.0} & \hi{21.0} & 17.1 \\
\midrule

\multicolumn{5}{l}{\textbf{Reference scope}} \\
International
& \en{35.0} & \bn{44.5} & \hi{63.5} & 47.7 \\
National
& \en{62.8} & \bn{54.3} & \hi{35.9} & 51.0 \\
Regional
& \en{2.1} & \bn{1.2} & \hi{0.6} & 1.3 \\
\bottomrule
\end{tabular}}

\caption{Difficulty-stratification composition of the full dataset across
\en{English (En)}, \bn{Bengali (Bn)}, and \hi{Hindi (Hi)}.
Dominant difficulty type and reference scope each sum to approximately
100\% per column due to rounding. The four ``property present'' rows are
independent Yes-rates and therefore do not sum.}
\label{tab:difficulty_stratification}
\vspace{-5mm}
\end{table}

\section{Experimental Setting}
\label{sec:setup}
With real-world deployment in mind, we evaluate a diverse set of VLMs spanning both closed and open-source families. On the closed-source side, we include several models from the GPT and Gemini Flash lineages, while the open-source selection covers a range of parameter scales across multiple model families. We also include Gemini 2.5 Pro as a reference point for its strong multilingual performance \cite{geminiteam2025gemini}.

\noindent \textbf{Evaluation Protocol:}
Each model was evaluated on Bengali, English, and Hindi subsets under two conditions: a \textbf{meme-only} setting, where the model received only the image, and a \textbf{context-aware} setting, where a brief cultural context note was additionally provided. As shown in Table~\ref{tab:explanation_comparison_multiple} (Appendix~\ref{sec:example_explanation_vlm}), even minimal context can shift a generic visual description toward a more culturally grounded interpretation. A fixed prompt (Appendix~\ref{sec:prompt-template}) instructed models to generate a single explanatory sentence, with temperature set to 0 and output capped at 100 tokens. Scores were averaged across three human references per meme. Full experimental details are provided in Appendix~\ref{sec:experimental-setup}.

\noindent \textbf{Automatic Metrics:}
Model-generated explanations were assessed against human references using three semantic metrics: \textbf{BERTScore-F1} \cite{zhang2020bertscore}, \textbf{SBERT similarity} \cite{reimers2019sentencebert}, and \textbf{BLEURT} \cite{sellam2020bleurt} chosen for their ability to capture semantic and contextual alignment beyond mere surface-level overlap.

\noindent \textbf{LLM-as-a-Judge and Human Evaluation:}
To complement automatic metrics, we employ Gemini 2.5 Pro as an LLM judge \cite{zheng2023judging} to score nine representative models on fluency, cultural correctness, and humor relevance using a 1--5 Likert scale. A parallel human
evaluation (N=3 raters, 100 memes per language) was conducted 
using the same rubric.

\section{Results and Analysis}
\label{sec:results}

\begin{table*}[t]
\centering
\small
\renewcommand{\arraystretch}{1.08}
\setlength{\tabcolsep}{2pt}
\resizebox{0.75\linewidth}{!}{
\begin{tabular}{lcc|c|cc|c}
\toprule
\textbf{Model} 
& \multicolumn{2}{c|}{\textbf{LLM as a Judge}} 
& \textbf{Avg. $\Delta$}
& \multicolumn{2}{c|}{\textbf{Human Evaluation}}
& \textbf{Avg. $\Delta$} \\
\cmidrule(lr){2-3} \cmidrule(lr){5-6}
& \textbf{Meme-only} & \textbf{Context-aware} 
&& \textbf{Meme-only} & \textbf{Context-aware}
& \\
\midrule

Gemini 2.5 Flash
& \en{3.70}\langsep\bn{3.84}\langsep\hi{4.03}
& \en{4.21}\langsep\bn{4.31}\langsep\hi{4.17}
& +0.37
& \en{3.90}\langsep\bn{3.75}\langsep\hi{4.00}
& \en{4.10}\langsep\bn{4.25}\langsep\hi{4.20}
& +0.30 \\

Gemini 2.5 Flash Lite
& \en{3.32}\langsep\bn{2.49}\langsep\hi{3.53}
& \en{4.02}\langsep\bn{3.74}\langsep\hi{3.91}
& +0.78
& \en{3.50}\langsep\bn{2.83}\langsep\hi{3.67}
& \en{3.91}\langsep\bn{3.95}\langsep\hi{3.98}
& +0.61 \\

GPT-5 Nano
& \en{3.04}\langsep\bn{1.98}\langsep\hi{3.37}
& \en{3.95}\langsep\bn{3.64}\langsep\hi{3.77}
& +0.99
& \en{3.45}\langsep\bn{2.40}\langsep\hi{3.67}
& \en{4.05}\langsep\bn{3.78}\langsep\hi{3.82}
& +0.71 \\

GPT-5 Mini
& \en{3.82}\langsep\bn{3.54}\langsep\hi{3.99}
& \en{4.36}\langsep\bn{4.33}\langsep\hi{4.26}
& +0.53
& \en{3.60}\langsep\bn{3.25}\langsep\hi{3.40}
& \en{4.20}\langsep\bn{4.00}\langsep\hi{3.80}
& +0.58 \\

\midrule

Gemma-3-12B
& \en{2.19}\langsep\bn{1.32}\langsep\hi{2.16}
& \en{2.87}\langsep\bn{2.74}\langsep\hi{2.82}
& +0.92
& \en{2.15}\langsep\bn{1.63}\langsep\hi{1.80}
& \en{3.00}\langsep\bn{2.63}\langsep\hi{2.25}
& +0.77 \\

InternVL3.5-2B
& \en{1.91}\langsep\bn{1.02}\langsep\hi{1.61}
& \en{3.04}\langsep\bn{2.15}\langsep\hi{2.64}
& +1.10
& \en{2.11}\langsep\bn{1.41}\langsep\hi{1.95}
& \en{3.23}\langsep\bn{2.75}\langsep\hi{2.96}
& +1.16 \\

InternVL3.5-8B
& \en{2.39}\langsep\bn{1.38}\langsep\hi{2.41}
& \en{3.33}\langsep\bn{3.03}\langsep\hi{3.31}
& +1.16
& \en{2.51}\langsep\bn{1.94}\langsep\hi{2.74}
& \en{3.64}\langsep\bn{3.45}\langsep\hi{3.56}
& +1.15 \\

Qwen3VL-2B
& \en{1.95}\langsep\bn{1.33}\langsep\hi{1.50}
& \en{3.05}\langsep\bn{2.31}\langsep\hi{2.49}
& +1.02
& \en{2.34}\langsep\bn{1.97}\langsep\hi{2.04}
& \en{3.24}\langsep\bn{3.05}\langsep\hi{2.50}
& +0.81 \\

Qwen3VL-8B
& \en{2.71}\langsep\bn{2.33}\langsep\hi{2.41}
& \en{3.60}\langsep\bn{3.25}\langsep\hi{3.30}
& +0.90
& \en{2.91}\langsep\bn{2.91}\langsep\hi{2.36}
& \en{3.96}\langsep\bn{3.78}\langsep\hi{3.65}
& +1.07 \\

\bottomrule
\end{tabular}}
\caption{\small{Results using LLM-as-a-Judge and Human Evaluation. Each entry reports \en{English} \langsep \bn{Bengali} \langsep \hi{Hindi}. Avg. $\Delta$ denotes the average improvement after providing context across all three languages. Higher is better.}}
\label{tab:judge_eval}
\vspace{-2mm}
\end{table*}

\subsection{Automatic Metric Evaluation}
Table~\ref{tab:quantitative_results} shows the performance of different VLMs on our multilingual meme dataset for SBERT Similarity, BERTScore F1, and BLEURT. 

\noindent \textbf{Effect of Context:} Cultural context consistently improves performance across all languages and metrics, though scores remain far from saturation, reflecting the inherent difficulty of meme understanding.

\noindent \textbf{Language-wise Trends:}
English consistently achieves the highest scores across all metrics and models, reflecting stronger cultural coverage during pretraining. Bengali consistently scores lowest overall but gains the most from added context, suggesting that meme text alone carries limited semantic signal and models rely heavily on explicit background cues.

\noindent \textbf{Model Comparison:} Among closed-source models, Gemini 2.5 Pro and Gemini 2.5 Flash Lite achieve the strongest results. Within open-source models, InternVL3.5 and Qwen3-VL families benefit most from context, with InternVL3.5-4B and Qwen3-VL-2B reaching up to 67.7 SBERT on English, surpassing several closed-source models. Notably, smaller variants often outperform larger ones, as InternVL3.5-4B surpasses InternVL3.5-8B across most context-aware metrics, suggesting parameter scale does not reliably improve context utilization. The Gemma family consistently underperforms across all metrics and languages. Bootstrap confidence intervals and the significance test results 
can be found in Appendix~\ref{app:significance}. 

\begin{table}[t]
\centering
\small
\setlength{\tabcolsep}{2pt}
\renewcommand{\arraystretch}{1.12}
\resizebox{\linewidth}{!}{
\begin{tabular}{l cc cc cc !{\vrule width 0.8pt} cc cc cc}
\toprule
& \multicolumn{6}{c!{\vrule width 0.8pt}}{\textbf{Gemini 2.5 Pro} (closed)}
& \multicolumn{6}{c}{\textbf{Gemma 12B} (open)} \\
\cmidrule(lr){2-7} \cmidrule(lr){8-13}

& \multicolumn{2}{c}{\en{English}}
& \multicolumn{2}{c}{\bn{Bengali}}
& \multicolumn{2}{c!{\vrule width 0.8pt}}{\hi{Hindi}}
& \multicolumn{2}{c}{\en{English}}
& \multicolumn{2}{c}{\bn{Bengali}}
& \multicolumn{2}{c}{\hi{Hindi}} \\
\cmidrule(lr){2-3} \cmidrule(lr){4-5} \cmidrule(lr){6-7}
\cmidrule(lr){8-9} \cmidrule(lr){10-11} \cmidrule(lr){12-13}

\textbf{Error Type} & E\% & R\% & E\% & R\% & E\% & R\%
                   & E\% & R\% & E\% & R\% & E\% & R\% \\
\midrule

Cultural Knowledge Gap
& \en{16.3} & \en{13.3} & \bn{26.3} & \bn{33.3} & \hi{16.7} & \hi{21.4}
& \en{34.0} & \en{9.4} & \bn{63.0} & \bn{22.2} & \hi{49.0} & \hi{18.7} \\

Entity Misidentification
& \en{41.3} & \en{26.3} & \bn{23.1} & \bn{52.4} & \hi{39.3} & \hi{12.1}
& \en{12.0} & \en{27.3} & \bn{7.0} & \bn{17.4} & \hi{8.0} & \hi{6.7} \\

Linguistic Failure
& \en{13.0} & \en{16.7} & \bn{19.8} & \bn{38.9} & \hi{15.5} & \hi{0.0}
& \en{21.0} & \en{5.3} & \bn{12.0} & \bn{15.0} & \hi{9.0} & \hi{11.8} \\

Literal Interpretation
& \en{9.8} & \en{22.2} & \bn{8.8} & \bn{50.0} & \hi{7.1} & \hi{0.0}
& \en{24.0} & \en{22.7} & \bn{12.0} & \bn{22.0} & \hi{22.0} & \hi{25.0} \\

Partial Humor Explanation
& \en{9.8} & \en{22.2} & \bn{18.7} & \bn{52.9} & \hi{8.3} & \hi{42.9}
& \en{3.0} & \en{33.3} & \bn{1.0} & \bn{66.7} & \hi{5.0} & \hi{20.0} \\

Hallucination
& \en{7.6} & \en{28.6} & \bn{2.2} & \bn{0.0} & \hi{11.9} & \hi{30.0}
& \en{5.0} & \en{20.0} & \bn{4.0} & \bn{35.7} & \hi{5.0} & \hi{33.3} \\

Visual Grounding Failure
& \en{2.2} & \en{0.0} & \bn{1.1} & \bn{100.0} & \hi{1.2} & \hi{100.0}
& \en{1.0} & \en{100.0} & \bn{1.0} & \bn{0.0} & \hi{2.0} & \hi{0.0} \\

\bottomrule
\end{tabular}
}
\caption{\small{Error analysis of meme-humor explanations from Gemini 2.5 Pro and Gemma 12B across \en{English}, \bn{Bengali}, and \hi{Hindi}. \textbf{E\%} shows error distribution; \textbf{R\%} shows context-based recovery per category.}}
\vspace{-3mm}
\label{tab:error-analysis-combined}
\end{table}

\subsection{Judge-Based Evaluation}
Table~\ref{tab:judge_eval} reports Likert scores (1--5) from both the LLM judge and human raters. Notably, we find that the LLM-judge scores correlated strongly with humans (Spearman $\rho = 0.92$).

\noindent \textbf{Context gap:} Judge-based evaluation reveals larger gaps between meme-only and context-aware settings than automatic metrics indicate. Gemma-3-12B in Bengali improves from (1.32$\to$2.74) under LLM-as-a-Judge and (1.63$\to$2.63) under human evaluation after adding context. GPT-5 Nano in Bengali rises from (1.98$\to$3.64) and (2.40$\to$3.78) respectively. These gains suggest contextual grounding yields qualitative improvements only partially reflected by surface-level metrics such as BERTScore or BLEURT.

\noindent \textbf{Open vs.\ closed source gap:} Judge-based evaluation highlights the open/closed-source disparity that automatic metrics often compress. In the meme-only setting, open-source models perform substantially worse, particularly in Bengali (Gemma-3-12B: 1.32; InternVL3.5-8B: 1.38 under LLM-as-a-Judge). Both improve notably with context (1.32$\to$2.74; 1.38$\to$3.03), suggesting these models can leverage external context effectively but struggle to infer 
cultural cues independently. To verify that these findings are not an artifact of judge–model family overlap, we further re-scored all models with \textbf{Qwen3-32B} in Appendix~\ref{sec:judge-bias}.

\subsection{Error Analysis}
\label{sec:error-analysis}

To complement aggregate metrics, we analyze failures of the best closed model, Gemini~2.5~Pro, and the largest open model, Gemma~12B, using seven error categories (full definitions in Appendix~\ref{app:error-categories}) and measuring whether added context resolves each error (Table~\ref{tab:error-analysis-combined}). The models fail in distinct ways. Gemini is primarily dominated by \emph{Entity Misidentification}, often understanding the meme structure but failing to identify the correct cultural entity or reference. In contrast, Gemma is dominated by \emph{Cultural Knowledge Gap} errors, particularly for \bn{Bengali} and \hi{Hindi}, indicating weaker regional cultural priors in open-source models. 

Across both models, \emph{Linguistic Failure} remains among the least recoverable categories, suggesting that short context notes are insufficient for resolving phonological, cross-script, or dialect-specific wordplay. \emph{Literal Interpretation} and \emph{Partial Humor Explanation} also remain common, especially in Bengali memes, where humor often depends on implicit cultural assumptions. Gemini generally benefits more consistently from context, whereas Gemma shows relatively weak recovery for \emph{Cultural Knowledge Gap} errors. Finally, \emph{Visual Grounding Failure} remains rare ($\leq$2.2\%), confirming that the primary bottleneck is cultural-linguistic reasoning rather than visual perception.


\subsection{Regional Meme Analysis}


\definecolor{regbn}{RGB}{190,45,45}      
\definecolor{regbnlight}{RGB}{255,235,235}
\definecolor{regbndark}{RGB}{120,20,20}

\newcommand{\reg}[1]{\cellcolor{regbnlight}\textcolor{regbndark}{#1}}
\newcommand{\regctx}[1]{\cellcolor{regbn!18}\textcolor{regbndark}{#1}}

\begin{table}[t]
\centering
\small
\renewcommand{\arraystretch}{1.0}
\setlength{\tabcolsep}{2pt}
\resizebox{\linewidth}{!}{
\begin{tabular}{lcc|cc|cc}
\toprule
\textbf{Model}
& \multicolumn{2}{c}{\textbf{SBERT Similarity}}
& \multicolumn{2}{c}{\textbf{BERTScore F1}}
& \multicolumn{2}{c}{\textbf{BLEURT}} \\
\cmidrule(lr){2-3} \cmidrule(lr){4-5} \cmidrule(lr){6-7}
& \textbf{Meme} & \textbf{CA}
& \textbf{Meme} & \textbf{CA}
& \textbf{Meme} & \textbf{CA} \\
\midrule

Gemini 2.5 Pro
& \reg{59.6} & \regctx{59.5}
& \reg{86.7} & \regctx{86.8}
& \reg{42.1} & \regctx{42.9} \\

Gemini 2.5 Flash
& \reg{52.8} & \regctx{55.9}
& \reg{85.7} & \regctx{86.3}
& \reg{36.0} & \regctx{39.1} \\

Gemini 2.5 Flash Lite
& \reg{42.4} & \regctx{51.6}
& \reg{84.8} & \regctx{85.5}
& \reg{33.6} & \regctx{37.0} \\

\bottomrule
\end{tabular}}
\caption{\small{Regional Bengali dialect meme results. Meme and CA denote the meme-only and context-aware settings, respectively.}}
\label{tab:regional_results}
\end{table}

Table~\ref{tab:regional_results} reports results on memes written in our additional Bengali regional dialectal variant subset. Compared to standard Bengali memes, these memes are substantially more challenging due to non-standard spellings, dialect-specific vocabulary, and localized humor conventions. Among the evaluated models, Gemini~2.5~Pro achieves the strongest overall performance in the meme-only setting, with only marginal gains from contextual information, suggesting robust regional-cultural priors even without explicit context. Context improves performance consistently across all models, with the largest gains in Gemini~2.5~Flash~Lite.

\section{Conclusion and Future Work}

We presented \textsc{MemeCULT-1K}, a multilingual benchmark for culturally grounded meme understanding in South Asian languages. Across thirteen VLMs, minimal cultural context consistently improves explanation quality across automatic metrics, LLM judging, and human evaluation, though overall performance remains far from ceiling. Error analysis shows distinct failure modes: closed-source models often misidentify cultural references, while open-source models lack regional background knowledge. Linguistic and phonological failures remain largely resistant to contextual cues, motivating future work on culturally grounded and cross-script multimodal reasoning.

\section*{Limitations}
Our dataset focuses on South Asian culture, specifically Bengali, Hindi,
and English memes, and findings may not generalize to other regions or
linguistic traditions. The benchmark is balanced across languages but modest in per-language scale, so results computed within a single language-metric cell or within a single error category should be read as indicative rather than definitive. Cultural humor is inherently subjective and
temporally fluid; interpretations grounded in current cultural references
may shift as memes and social conventions evolve. Automatic metrics such
as BERTScore, SBERT, and BLEURT approximate semantic similarity but do
not fully capture humor understanding, irony, or pragmatic nuance, which
is why we complement them with LLM-as-a-judge and human evaluation. The
LLM judge evaluation covers a representative subset of models selected
to span both open and closed-source families. Our cultural context notes
are concise by design, which may limit their ability to fully resolve
deeply entrenched cultural or phonological wordplay, as also observed in
our error analysis. Future work will expand to additional South Asian
languages such as Tamil, Urdu, and Marathi, explore retrieval-augmented
context grounding, and conduct larger-scale human evaluation across
diverse annotator populations.

\section*{Ethics Statement}
All memes were publicly available and screened to remove NSFW or offensive content. While efforts were made to minimize bias, humor interpretation remains subjective and culturally situated, and explanations may reflect annotator perspectives. Users should exercise caution in downstream applications. All participating annotators were paid well above their country-specific minimum wages. 

\section*{Acknowledgements}
We thank Islamic University of Technology for funding the
annotation effort, CUPE 3903 Research Grant Fund at York University for providing API credits, Google for Gemini API credits, and the Digital Research Alliance of Canada for computing resources. We are also grateful to the anonymous reviewers for their constructive feedback, which substantially improved this work.

\bibliography{custom}

\appendix
\section{Appendix}
\label{sec:appendix}

\begin{table}[h!]
\centering
\small
\renewcommand{\arraystretch}{1.15}
\setlength{\tabcolsep}{3pt}

\begin{tabular}{p{0.25\linewidth}p{0.67\linewidth}}
\toprule
\textbf{Filename} & English\_0001.jpg \\
\textbf{Language} & English \\
\textbf{Context} & The meme is about Mushfiqur Rahim and Nurul Hasan Shohan making similar cricketing mistakes. \\
\textbf{Explanation\_1} & The meme contextualizes how both Mushfiqur Rahim and Nurul Hasan Shohan commit similar cricketing errors by breaking the stump before catching the ball. \\
\textbf{Explanation\_2} & This meme humorously points out that both Mushfiqur Rahim and Nurul Hasan Shohan make the same mistake, breaking the stump before completing the catch. \\
\textbf{Explanation\_3} & The image exaggerates the comedic similarity in their blunders, turning a cricket mishap into a light-hearted joke. \\
\textbf{Annotator IDs} & A03, A05, A09 \\
\bottomrule
\end{tabular}
\caption{Representative example entry from the \textsc{MemeCULT-1K} dataset (CSV format). Values are illustrative.}
\label{tab:annotation_csv_vertical}
\end{table}

\begin{table*}[t]
\centering
\small
\renewcommand{\arraystretch}{1.2}

\resizebox{1.9\columnwidth}{!}{
\begin{tabularx}{\textwidth}{
  >{\raggedright\arraybackslash}X
  >{\raggedright\arraybackslash}X
  >{\raggedright\arraybackslash}X
}
\toprule

\makecell[c]{\includegraphics[width=0.9\linewidth]{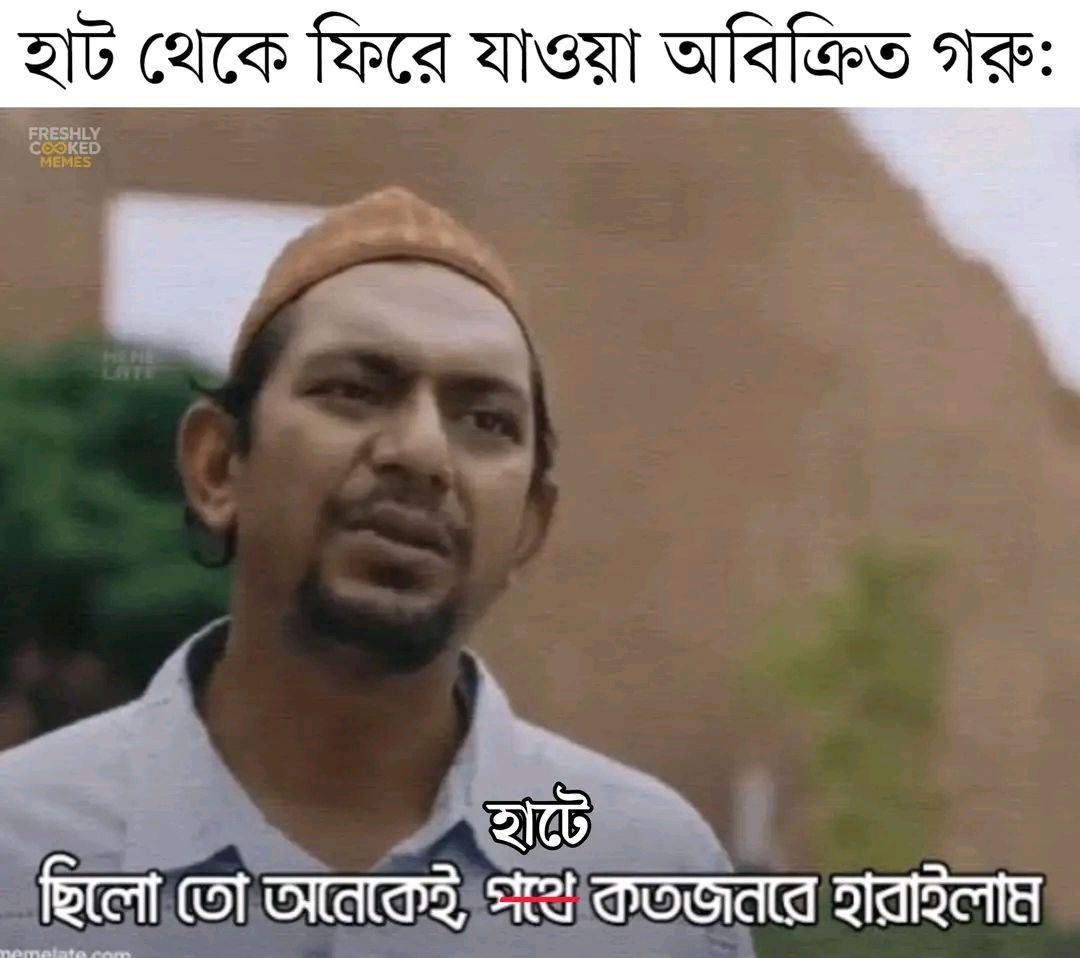}} &
\makecell[c]{\includegraphics[width=0.9\linewidth]{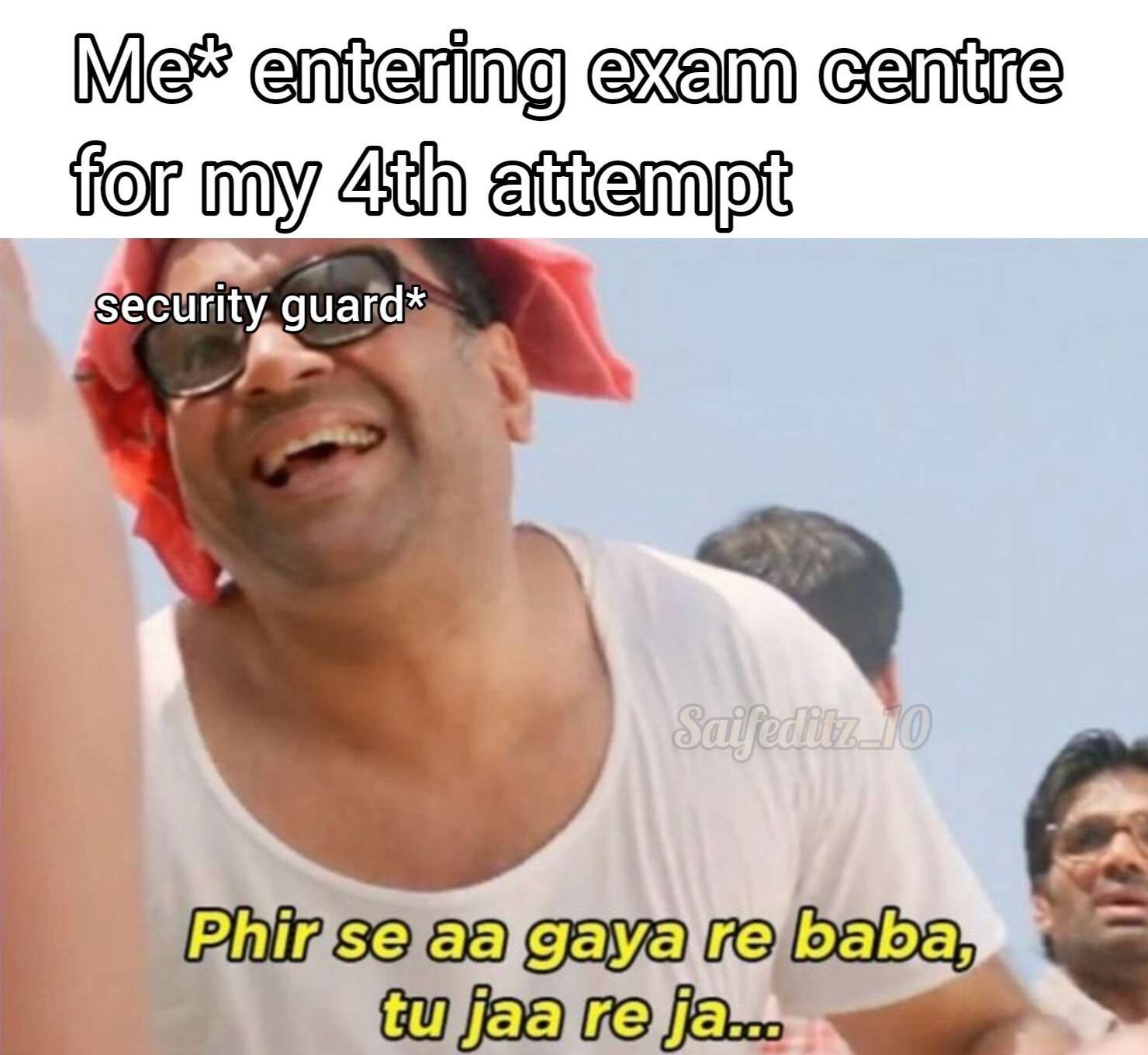}} &
\makecell[c]{\includegraphics[width=0.8\linewidth]{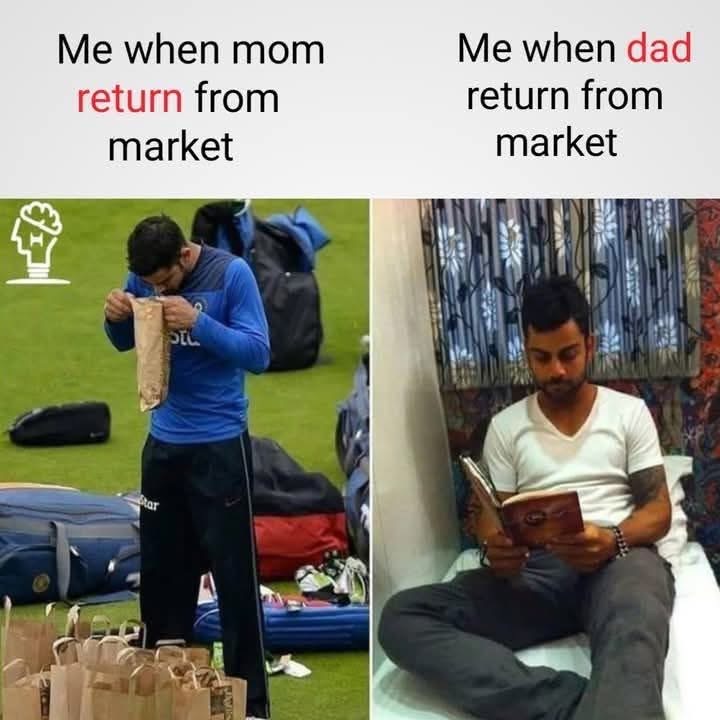}} \\

\midrule

\textbf{Context:} Depicts unsold cows at the Qurbani market after the festival ends. &
\textbf{Context:} Depicts the repeated interaction between a failing student and an exam guard. &
\textbf{Context:} Contrasts children’s reactions to their mother’s versus father’s return from the market. \\

\midrule

\textbf{Explanation:} Shows the sorrow of cows that return unsold, using emotional irony to anthropomorphize livestock. &
\textbf{Explanation:} The guard mocks the student’s persistence, symbolizing repeated academic failure humorously. &
\textbf{Explanation:} Children cheer when their mother arrives expecting snacks but act studious when their father returns home. \\

\bottomrule

\end{tabularx}}
\caption{Example memes with corresponding context notes and human explanations.}

\label{tab:example_meme_context_explanation}
\end{table*}

\subsection{Annotation Guidelines}
\label{sec:annotation-guidelines}
13 Annotators were hired among university students based on their urban culture and meme knowledge. They were hired in contract and were paid in full upon the completion of the annotation work. To ensure consistent and culturally sensitive meme explanations, all annotators followed a standardized set of guidelines. Each meme was annotated independently by three annotators, fluent in English and familiar with at least one cultural region represented in the dataset (Bengali, Hindi, or English).

\paragraph{General Instructions.}
Annotators were instructed to provide short, objective explanations that clearly described each meme’s meaning or humor while maintaining a neutral tone. They were asked to avoid reproducing or quoting any potentially offensive text verbatim, ensure grammatical correctness, and remain culturally balanced in their phrasing. Personal commentary, subjective humor, or political interpretation was explicitly discouraged to maintain consistency and fairness across all samples.

\paragraph{Explanation Writing Protocol.}
Each meme was annotated with three distinct English explanations (\texttt{exp1–3}), written independently by different annotators. While the three versions preserved the same core interpretation, they varied in style and phrasing to capture natural linguistic diversity. Annotators were instructed to focus on accurately describing the visual–textual relationship within the meme and to include only as much cultural context as necessary for a reader unfamiliar with the source culture to grasp the intended humor or message.

\paragraph{Cultural Context Notes.}
Each meme was accompanied by a short (2-3 sentences) context note summarizing cultural or situational background. Annotators were encouraged to identify cultural markers (e.g., cricket, film tropes, or local idioms) influencing humor.

\paragraph{Annotation Format.}
Each record in \textsc{MemeCULT-1K} corresponds to a single meme image stored as a CSV row. 
Columns include the meme filename, cultural context note, three English explanations written by independent annotators, and their anonymized IDs. 
Table~\ref{tab:annotation_csv_vertical} shows one representative example from the dataset, formatted vertically for readability.

\paragraph{Inter-Annotator Agreement.}
Since explanations are open-ended natural language texts rather than categorical labels, inter-annotator consistency was assessed using pairwise SBERT cosine similarity between the three explanations per meme. The mean similarity score across all samples was 0.78, indicating strong semantic agreement in annotators’ interpretations.

\subsection{Example Annotations and Contexts}
\label{sec:example-annotations}

Table~\ref{tab:example_meme_context_explanation} presents representative samples from the Bengali, Hindi, and English subsets. All images are resized to fit within column width and anonymized for ethical compliance.

\subsection{Example explanations}
\label{sec:example_explanation_vlm}

Example explanations for memes generated by VLMs are provided in Table \ref{tab:explanation_comparison_multiple}.

\begin{table*}[t]
\centering
\renewcommand{\arraystretch}{1.25}
\resizebox{1.9\columnwidth}{!}{
\begin{tabular}{|c|p{0.65\textwidth}|}
\hline
\textbf{Meme Image} & \textbf{Explanation} \\ \hline

\multirow{3}{*}{%
  \begin{minipage}[c]{0.25\linewidth}
    \includegraphics[width=0.92\linewidth]{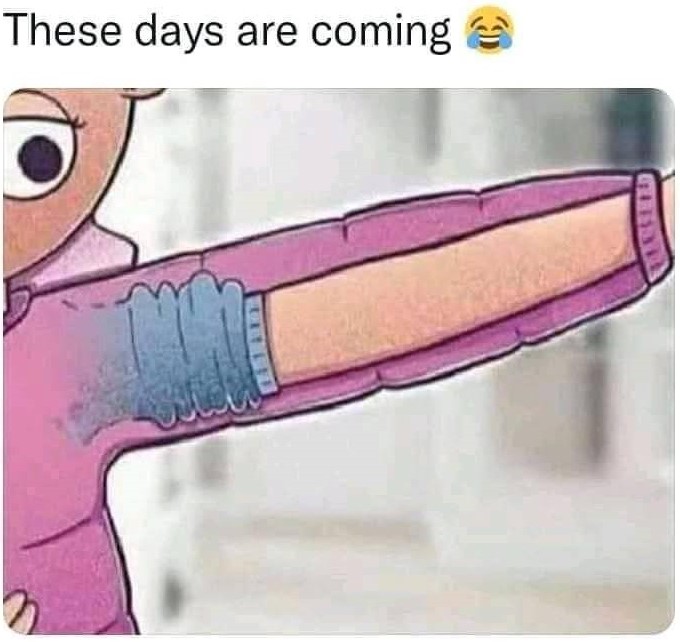}
  \end{minipage}
}
& \scriptsize\textbf{Ground Truth:} This meme humorously depicts the common struggle of trying to put on a second long-sleeved shirt over a first one to combat cold weather. The sleeve of the inner shirt often gets bunched up and pulled back towards the elbow, causing frustration. \\ \cline{2-2}
& \scriptsize\textbf{Meme-only:} It shows a cartoon person wearing a long-sleeved sweater, with the sleeve pulled up, revealing sweat stains around the armpit area. Basically, it’s a relatable joke about sweating in sweaters when the season shifts. \\ \cline{2-2}
& \scriptsize\textbf{Context-aware:} This meme humorously highlights the everyday struggle of layering clothes during cold weather. When someone tries to put on a second long-sleeved shirt or sweater over the first one, the inner sleeve often gets dragged upward and bunched around the elbow, making it both uncomfortable and frustrating. \\ \hline

\multirow{3}{*}{%
  \begin{minipage}[c]{0.25\linewidth}
    \includegraphics[width=0.95\linewidth]{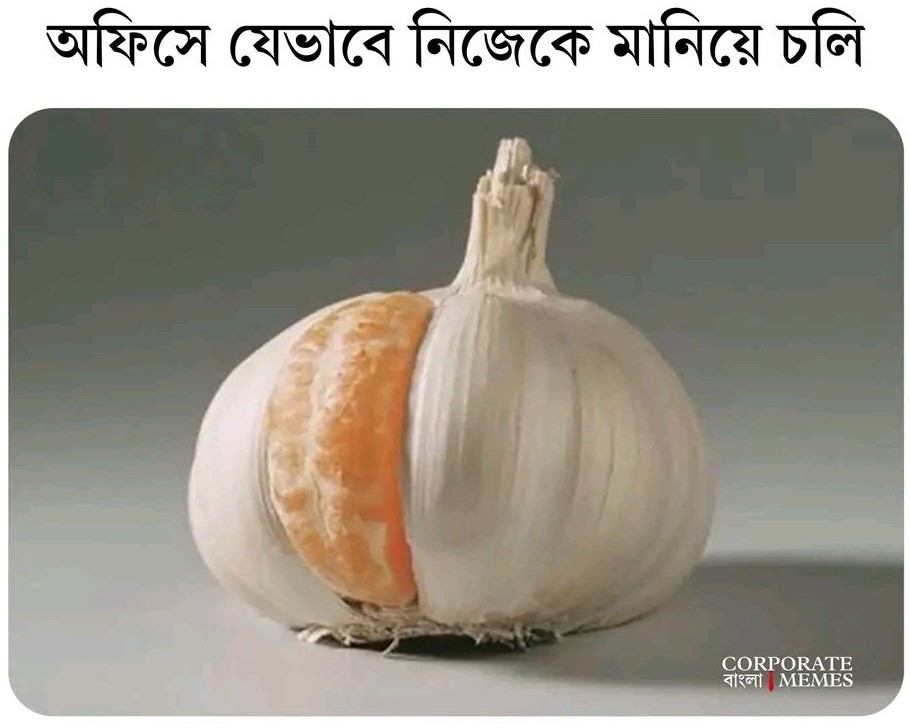}
  \end{minipage}
}
& \scriptsize\textbf{Ground Truth:} The meme is about how an employee tries to fit in his/her office even though they have differences with the rest of the team. The image hilariously portrays the contrast with a slice of orange trying to fit into a garlic. \\ \cline{2-2}
& \scriptsize\textbf{Meme-only:} Garlic is known for its strong, pungent flavor and smell, and its tough, papery outer skin. In this metaphor, the garlic represents the tough, often demanding, or perhaps dull external environment of the office. \\ \cline{2-2}
& \scriptsize\textbf{Context-aware:} The meme signifies that the employee is masking their authentic self and forcing themselves to fit a rigid professional mold, showing only a tough, neutral shell to their colleagues while keeping their true nature hidden inside to avoid conflict or stand out. \\ \hline

\end{tabular}}
\caption{\small{Example explanations for memes generated by VLMs under meme-only vs.\ context-aware settings.}}
\label{tab:explanation_comparison_multiple}
\end{table*}

\subsection{VLM Evaluation Prompt}
\label{sec:prompt-template}

For consistent automated evaluation, all vision–language models were prompted with the following standardized instruction:

{\ttfamily
Write one clear and self-contained sentence that explains the meaning or humor expressed in the content. 
Focus only on the underlying idea or irony implied. 
Avoid using words such as "meme", "image", "picture", "joke", or "pun". 
Do not describe visual details or add commentary. 
Return only the explanatory sentence.
}

\subsection{Experimental Setup}
\label{sec:experimental-setup}
For each meme, we use a fixed instruction prompt. We send the image as a base64-encoded input, prepend the context note only in the context-aware condition, and use deterministic decoding (temperature $=0$; outputs limited to roughly 100 tokens). For each item, model outputs are compared against three independent reference explanations, and automatic scores are computed by averaging scores over references. We run this pipeline separately for each language split and save per-meme outputs and scores for analysis.

\subsection{Cross-Judge Robustness of Evaluation}
\label{sec:judge-bias}

\begin{table}[h]
\centering
\small
\renewcommand{\arraystretch}{1.08}
\setlength{\tabcolsep}{3pt}
\resizebox{0.85\columnwidth}{!}{
\begin{tabular}{lcc|cc|cc}
\toprule
\textbf{Model}
& \multicolumn{2}{c|}{\textbf{Human}}
& \multicolumn{2}{c|}{\textbf{Gemini 2.5 Pro}}
& \multicolumn{2}{c}{\textbf{Qwen3-32B}} \\
\cmidrule(lr){2-3} \cmidrule(lr){4-5} \cmidrule(lr){6-7}
& \textbf{M} & \textbf{C}
& \textbf{M} & \textbf{C}
& \textbf{M} & \textbf{C} \\
\midrule

Gemini 2.5 Flash       & 3.88 & 4.18 & 3.86 & 4.23 & 3.42 & 3.68 \\
GPT-5 Mini             & 3.42 & 4.00 & 3.78 & 4.32 & 3.45 & 3.85 \\
Gemini 2.5 Flash Lite  & 3.33 & 3.95 & 3.11 & 3.89 & 3.05 & 3.58 \\
GPT-5 Nano             & 3.17 & 3.88 & 2.80 & 3.79 & 2.83 & 3.53 \\

\midrule

Qwen3-VL-8B            & 2.73 & 3.80 & 2.48 & 3.38 & 2.30 & 3.27 \\
InternVL3.5-8B         & 2.40 & 3.55 & 2.06 & 3.22 & 2.16 & 3.16 \\
InternVL3.5-2B         & 1.82 & 2.98 & 1.51 & 2.61 & 1.88 & 2.99 \\
Qwen3-VL-2B            & 2.12 & 2.93 & 1.59 & 2.62 & 1.90 & 2.93 \\
Gemma-3-12B            & 1.86 & 2.63 & 1.89 & 2.81 & 2.13 & 2.84 \\

\bottomrule
\end{tabular}}
\caption{\small{Cross-judge robustness. Scores (1--5) averaged over
English/Bengali/Hindi. \textbf{M}/\textbf{C} denote meme-only and
context-aware settings. Human evaluation uses 100 memes per language (as in
the main paper); LLM judges score the full subsets.}}
\label{tab:judge_bias}
\vspace{-3mm}
\end{table}

Because our LLM judge (Gemini~2.5~Pro) shares a model family with several
evaluated systems, one may worry that judge scores are inflated by
self-preference. To test this, we re-scored all models with
\textbf{Qwen3-32B}, an open-weight judge from a different vendor, using the
identical prompt and rubric as Section~\ref{sec:setup}. We note that the
Gemini~2.5~Pro judge never scored its own outputs.
Table~\ref{tab:judge_bias} reports human evaluation alongside both LLM judges,
with all scores averaged over English, Bengali, and Hindi.

\paragraph{Human evaluation corroborates the Gemini judge.}
If the judge favored its own family, human ratings should diverge from it.
Instead, they agree within $0.06$ points on the Gemini systems in the
context-aware setting (Flash: $4.18$ human vs.\ $4.23$ judge; Flash~Lite:
$3.95$ vs.\ $3.89$), indicating that the high Gemini scores reflect
independent human judgment rather than self-preference.

\paragraph{Family overlap does not inflate scores.}
Qwen3-32B likewise judges its own vendor's models yet shows no favoritism,
scoring Qwen3-VL-8B at $3.27$ versus $3.80$ (human) and $3.38$ (Gemini judge).
Its uniformly lower scores apply broadly across the Gemini, GPT-5, and Qwen
families alike, reflecting stricter grading rather than family-directed bias.

\paragraph{Conclusions are evaluator-invariant.}
Context gains are positive for every model under all three evaluators, and
model rankings agree strongly (Spearman $\rho$ over the nine common models:
human--Qwen3-32B $0.98$; human--Gemini judge $0.92$; between judges $0.93$).
No conclusion in the paper depends on absolute judge scores; accordingly, LLM
judge scores should be read as relative, not absolute, quality measures.

\subsection{Statistical Significance and Confidence Intervals}
\label{app:significance}

To quantify variance and significance, we report 95\% percentile bootstrap
confidence intervals (10{,}000 resamples over memes) for every cell across all
three metrics, and test each context gain with a paired bootstrap test under
Holm--Bonferroni correction in Table~\ref{tab:significance}. 
\begin{table*}[h]
\centering
\footnotesize
\setlength{\tabcolsep}{4pt}
\resizebox{\textwidth}{!}{%
\begin{tabular}{ll ccc ccc ccc}
\toprule
& & \multicolumn{3}{c}{\textbf{SBERT}} & \multicolumn{3}{c}{\textbf{BERTScore F1}} & \multicolumn{3}{c}{\textbf{BLEURT}} \\
\cmidrule(lr){3-5}\cmidrule(lr){6-8}\cmidrule(lr){9-11}
\textbf{Model} & \textbf{Lang} & Meme-only & Context & $\Delta$ & Meme-only & Context & $\Delta$ & Meme-only & Context & $\Delta$ \\
\midrule
\multirow{3}{*}{GPT-5 Mini} & EN & 51.9 [50.7, 53.2] & 56.0 [55.0, 57.0] & +4.1 & 85.7 [85.5, 85.9] & 86.1 [85.9, 86.3] & +0.4 & 40.2 [39.6, 40.9] & 41.3 [40.7, 41.9] & +1.1 \\
 & BN & 49.7 [48.2, 51.1] & 58.8 [57.7, 59.8] & +9.1 & 85.1 [84.9, 85.3] & 85.9 [85.7, 86.1] & +0.8 & 39.1 [38.4, 39.8] & 42.3 [41.7, 42.9] & +3.2 \\
 & HI & 51.3 [50.0, 52.6] & 54.4 [53.2, 55.6] & +3.1 & 85.3 [85.1, 85.5] & 85.5 [85.3, 85.7] & +0.2 & 39.7 [39.0, 40.3] & 41.1 [40.5, 41.7] & +1.4 \\
\midrule
\multirow{3}{*}{GPT-5 Nano} & EN & 48.1 [46.7, 49.4] & 54.8 [53.6, 55.9] & +6.7 & 85.6 [85.4, 85.8] & 86.2 [86.0, 86.4] & +0.6 & 39.1 [38.5, 39.7] & 41.2 [40.5, 41.8] & +2.1 \\
 & BN & 36.8 [35.3, 38.4] & 55.3 [54.1, 56.5] & +18.5 & 84.3 [84.1, 84.5] & 85.7 [85.5, 85.9] & +1.4 & 33.1 [32.5, 33.7] & 40.7 [40.1, 41.3] & +7.6 \\
 & HI & 49.3 [48.0, 50.6] & 52.7 [51.5, 53.9] & +3.4 & 85.1 [84.9, 85.3] & 85.5 [85.3, 85.7] & +0.4 \textit{(n.s.)} & 38.7 [38.1, 39.3] & 40.6 [40.0, 41.2] & +1.9 \\
\midrule
\multirow{3}{*}{Gemini 2.5 Pro} & EN & 57.1 [56.0, 58.2] & 59.4 [58.3, 60.5] & +2.3 & 86.9 [86.7, 87.1] & 87.3 [87.1, 87.5] & +0.4 & 43.7 [43.0, 44.4] & 45.0 [44.4, 45.6] & +1.3 \\
 & BN & 56.9 [55.8, 58.0] & 58.6 [57.5, 59.7] & +1.7 & 86.7 [86.5, 86.9] & 86.9 [86.7, 87.1] & +0.2 & 44.0 [43.3, 44.7] & 45.2 [44.5, 45.9] & +1.2 \\
 & HI & 54.2 [52.8, 55.6] & 54.7 [53.5, 55.9] & +0.5 \textit{(n.s.)} & 86.2 [86.0, 86.4] & 86.4 [86.2, 86.6] & +0.2 \textit{(n.s.)} & 42.1 [41.5, 42.7] & 42.3 [41.6, 43.0] & +0.2 \textit{(n.s.)} \\
\midrule
\multirow{3}{*}{Gemini 2.5 Flash} & EN & 55.8 [54.5, 57.1] & 58.2 [57.0, 59.4] & +2.4 \textit{(n.s.)} & 86.5 [86.3, 86.7] & 86.8 [86.6, 87.0] & +0.3 & 41.8 [41.1, 42.5] & 43.2 [42.5, 43.9] & +1.4 \\
 & BN & 54.1 [52.9, 55.3] & 56.6 [55.5, 57.7] & +2.5 & 86.0 [85.8, 86.2] & 86.3 [86.1, 86.5] & +0.3 & 41.1 [40.4, 41.8] & 42.6 [41.9, 43.2] & +1.5 \\
 & HI & 53.8 [52.5, 55.1] & 54.9 [53.6, 56.1] & +1.1 \textit{(n.s.)} & 85.9 [85.7, 86.1] & 86.0 [85.8, 86.2] & +0.1 \textit{(n.s.)} & 40.9 [40.3, 41.5] & 41.9 [41.3, 42.5] & +1.0 \\
\midrule
\multirow{3}{*}{Gemini 2.5 Flash-Lite} & EN & 57.9 [56.7, 59.1] & 59.6 [58.4, 60.8] & +1.7 & 86.5 [86.3, 86.7] & 86.9 [86.7, 87.1] & +0.4 & 41.0 [40.3, 41.7] & 42.9 [42.3, 43.6] & +1.9 \\
 & BN & 51.8 [50.5, 53.1] & 59.1 [58.0, 60.1] & +7.3 & 85.5 [85.3, 85.7] & 86.5 [86.3, 86.7] & +1.0 & 38.7 [38.0, 39.3] & 41.1 [40.5, 41.7] & +2.4 \\
 & HI & 56.7 [55.4, 58.0] & 58.7 [57.6, 59.8] & +2.0 & 86.0 [85.8, 86.2] & 86.3 [86.1, 86.5] & +0.3 \textit{(n.s.)} & 41.5 [40.9, 42.1] & 42.9 [42.3, 43.5] & +1.4 \\
\midrule
\multirow{3}{*}{Gemma 3 4B} & EN & 31.6 [30.4, 32.8] & 37.6 [36.5, 38.7] & +6.0 & 85.6 [85.5, 85.7] & 85.8 [85.7, 85.9] & +0.2 & 33.9 [33.3, 34.5] & 36.1 [35.5, 36.7] & +2.2 \\
 & BN & 22.9 [21.9, 23.9] & 36.0 [34.9, 37.1] & +13.1 & 84.9 [84.8, 85.0] & 85.4 [85.3, 85.5] & +0.5 & 30.6 [30.1, 31.1] & 35.5 [34.9, 36.1] & +4.9 \\
 & HI & 31.4 [30.3, 32.5] & 36.9 [35.8, 38.0] & +5.5 & 85.4 [85.3, 85.5] & 85.5 [85.3, 85.7] & +0.1 \textit{(n.s.)} & 34.3 [33.7, 34.9] & 36.4 [35.8, 37.0] & +2.1 \\
\midrule
\multirow{3}{*}{Gemma 3 12B} & EN & 37.0 [35.9, 38.2] & 52.9 [51.8, 54.1] & +15.9 & 85.8 [85.6, 86.0] & 86.7 [86.5, 86.9] & +0.9 & 34.6 [34.0, 35.2] & 40.3 [39.7, 40.9] & +5.7 \\
 & BN & 24.8 [23.8, 25.8] & 49.5 [48.3, 50.7] & +24.7 & 84.7 [84.5, 84.9] & 85.9 [85.7, 86.1] & +1.2 & 28.4 [27.9, 28.9] & 38.0 [37.4, 38.6] & +9.6 \\
 & HI & 34.4 [33.3, 35.5] & 49.9 [48.8, 51.1] & +15.5 & 85.5 [85.3, 85.7] & 86.0 [85.8, 86.2] & +0.5 & 33.4 [32.8, 34.0] & 39.1 [38.5, 39.7] & +5.7 \\
\midrule
\multirow{3}{*}{InternVL 2B} & EN & 49.4 [47.9, 50.9] & 66.1 [64.8, 67.4] & +16.7 & 86.6 [86.4, 86.8] & 88.0 [87.8, 88.2] & +1.4 & 38.5 [37.8, 39.2] & 46.7 [45.9, 47.5] & +8.2 \\
 & BN & 25.3 [23.9, 26.7] & 62.2 [61.0, 63.4] & +36.9 & 84.9 [84.7, 85.1] & 87.2 [87.0, 87.4] & +2.3 & 31.0 [30.4, 31.6] & 44.3 [43.5, 45.1] & +13.3 \\
 & HI & 44.0 [42.5, 45.5] & 62.5 [61.2, 63.8] & +18.5 & 85.3 [85.1, 85.5] & 86.8 [86.6, 87.0] & +1.5 & 37.2 [36.5, 37.9] & 44.4 [43.7, 45.1] & +7.2 \\
\midrule
\multirow{3}{*}{InternVL 4B} & EN & 53.7 [52.3, 55.1] & 67.7 [66.6, 68.8] & +14.0 & 86.9 [86.7, 87.1] & 88.2 [88.0, 88.4] & +1.3 & 40.0 [39.3, 40.7] & 46.6 [45.8, 47.4] & +6.6 \\
 & BN & 28.6 [27.1, 30.1] & 63.4 [62.2, 64.6] & +34.8 & 84.9 [84.8, 85.0] & 87.2 [87.0, 87.4] & +2.3 & 31.2 [30.6, 31.8] & 44.0 [43.3, 44.7] & +12.8 \\
 & HI & 49.4 [48.0, 50.8] & 63.3 [62.1, 64.5] & +13.9 & 85.3 [85.1, 85.5] & 87.8 [87.5, 88.1] & +2.5 & 38.6 [37.9, 39.3] & 44.3 [43.6, 45.0] & +5.7 \\
\midrule
\multirow{3}{*}{InternVL 8B} & EN & 52.7 [51.4, 54.0] & 61.7 [60.6, 62.9] & +9.0 & 86.7 [86.5, 86.8] & 87.5 [87.4, 87.7] & +0.8 & 40.2 [39.5, 40.9] & 44.4 [43.7, 45.1] & +4.2 \\
 & BN & 30.8 [29.6, 32.1] & 58.2 [57.0, 59.4] & +27.4 & 85.0 [84.9, 85.1] & 86.8 [86.6, 86.9] & +1.8 & 31.1 [30.6, 31.7] & 42.3 [41.6, 43.0] & +11.2 \\
 & HI & 51.4 [50.0, 52.8] & 57.7 [56.4, 58.9] & +6.3 & 83.8 [83.6, 84.0] & 86.9 [86.7, 87.1] & +3.1 & 39.7 [39.0, 40.3] & 43.0 [42.3, 43.6] & +3.3 \\
\midrule
\multirow{3}{*}{Qwen3-VL 2B} & EN & 53.6 [52.3, 54.9] & 67.7 [66.6, 68.8] & +14.1 & 86.4 [86.3, 86.5] & 87.8 [87.7, 87.9] & +1.4 & 39.8 [39.2, 40.4] & 46.7 [46.0, 47.4] & +6.9 \\
 & BN & 31.7 [30.3, 33.1] & 62.8 [61.7, 63.9] & +31.1 & 84.3 [84.2, 84.4] & 87.0 [86.9, 87.1] & +2.7 & 32.3 [31.7, 32.9] & 43.9 [43.2, 44.6] & +11.6 \\
 & HI & 46.8 [45.3, 48.3] & 61.6 [60.3, 62.9] & +14.8 & 84.0 [83.8, 84.2] & 85.0 [84.8, 85.2] & +1.0 & 37.4 [36.8, 38.0] & 44.2 [43.6, 44.8] & +6.8 \\
\midrule
\multirow{3}{*}{Qwen3-VL 4B} & EN & 47.9 [46.5, 49.3] & 56.4 [55.2, 57.6] & +8.5 & 86.0 [85.8, 86.2] & 86.9 [86.7, 87.1] & +0.9 & 37.9 [37.3, 38.5] & 42.3 [41.6, 43.0] & +4.4 \\
 & BN & 28.0 [26.7, 29.3] & 55.5 [54.2, 56.8] & +27.5 & 84.7 [84.6, 84.8] & 86.4 [86.2, 86.6] & +1.7 & 30.3 [29.7, 30.9] & 41.4 [40.7, 42.1] & +11.1 \\
 & HI & 45.3 [43.9, 46.7] & 53.8 [52.4, 55.2] & +8.5 & 83.9 [83.8, 84.0] & 86.4 [86.2, 86.6] & +2.5 & 36.8 [36.1, 37.5] & 41.3 [40.7, 41.9] & +4.5 \\
\midrule
\multirow{3}{*}{Qwen3-VL 8B} & EN & 50.2 [48.7, 51.6] & 57.0 [55.8, 58.3] & +6.8 & 86.2 [86.0, 86.4] & 86.9 [86.7, 87.1] & +0.7 & 39.0 [38.3, 39.7] & 42.4 [41.7, 43.1] & +3.4 \\
 & BN & 34.1 [32.7, 35.5] & 56.8 [55.5, 58.0] & +22.7 & 84.9 [84.7, 85.1] & 86.4 [86.2, 86.6] & +1.5 & 32.0 [31.4, 32.6] & 41.3 [40.7, 42.0] & +9.3 \\
 & HI & 49.5 [48.2, 50.8] & 49.7 [48.4, 51.0] & +0.2 & 84.1 [83.9, 84.3] & 88.3 [88.1, 88.5] & +4.2 \textit{(n.s.)} & 38.2 [37.5, 38.9] & 40.8 [40.2, 41.5] & +2.6 \\
\bottomrule
\end{tabular}}
\caption{Per-language results for all 13 models under meme-only and context-aware settings, on SBERT similarity, BERTScore F1, and BLEURT (all $\times$100). Each cell reports the mean with its 95\% percentile-bootstrap confidence interval (10{,}000 resamples over memes). $\Delta$ is the mean paired per-meme context gain (context-aware $-$ meme-only); gains are significant at the 95\% level under a paired bootstrap test with Holm--Bonferroni correction within each metric family unless marked \textit{(n.s.)}.}
\label{tab:significance}
\end{table*}

\subsection{Error Category Definitions}
\label{app:error-categories}

For the error analysis in Section~5.3, two annotators independently
labeled each incorrect or incomplete model output into one of seven
mutually exclusive categories. When multiple error types co-occurred,
annotators selected the most salient one- the error type most directly
responsible for the explanation failing to match the ground truth.
Disagreements were resolved through discussion. Definitions and
representative examples are provided below.

\paragraph{Cultural / Social Knowledge Gap.}
The model fails to recognize the broader cultural, social, or
situational background required to interpret the meme. The model
correctly perceives the visual and textual content but cannot connect
it to the relevant cultural reference (e.g., a regional festival, a
local custom, or a social trope). Example: failing to recognize that
``Qurbani'' refers to the Eid al-Adha sacrificial festival.

\paragraph{Entity / Reference Misidentification.}
The model identifies that a specific entity is being referenced but
attributes it to the wrong person, place, event, or meme template.
Distinguished from Cultural Knowledge Gap in that the model recognizes
the \emph{kind} of reference but gets the specific anchor wrong.
Example: identifying a cricketer in the meme but naming the wrong
player.

\paragraph{Linguistic Failure.}
The model fails to correctly parse the meme's text due to code-mixing,
non-standard orthography, cross-script wordplay, or
phonological/transliteration patterns. Example: misreading romanized
Bengali (``Banglish'') as English, or failing to recognize a Hindi pun
that depends on Devanagari–Roman script switching.

\paragraph{Surface-Level / Literal Interpretation.}
The model describes only the visible content of the meme without
inferring its intended humorous or ironic meaning. Output is factually
correct but pragmatically empty (e.g., describing the meme as ``a man
holding a phone'' without noting the irony).

\paragraph{Incomplete Humor Explanation.}
The model partially captures the humor but misses a critical component,
such as the punchline, the cultural twist, or the subject of the joke.
Distinguished from Surface-Level in that some humorous intent is
recognized but the explanation is incomplete.

\paragraph{Hallucination / Fabricated Content.}
The model introduces information not present in the meme or its context,
such as inventing names, events, or backstories that do not exist in
the source. Includes confidently asserting incorrect cultural facts.

\paragraph{Multimodal Grounding Failure.}
The model's explanation contradicts or ignores clear visual evidence in
the meme (e.g., describing a person as smiling when they are crying, or
referring to objects not present in the image). Distinguished from
Hallucination in that the failure is specifically a vision–text
misalignment rather than fabricated world knowledge.

\subsection{Ethical Considerations}
\label{sec:ethics}
All memes were sourced from publicly available social media platforms
(Reddit, Facebook, and regional meme pages) for non-commercial research
purposes. NSFW or sensitive material was filtered in multiple stages
during dataset creation and annotation. Hateful and potentially harmful
memes were also removed. The dataset will be released for academic and
non-commercial research use only; original meme creators retain
copyright over the underlying images.

\end{document}